\documentclass[letterpaper, 10 pt, conference]{ieeeconf}  

\IEEEoverridecommandlockouts                              

\usepackage{xcolor}
\usepackage{comment}
\usepackage{paralist}
\usepackage{hyperref}
\usepackage{graphicx}
\usepackage{amsfonts}
\usepackage{amsmath}
\usepackage[english]{babel}
\usepackage{booktabs}
\usepackage{multirow}
\usepackage{subcaption}
\usepackage{floatrow}
\usepackage{sidecap}
\usepackage[affil-it]{authblk}
\usepackage[T1]{fontenc}
\usepackage[scientific-notation=true]{siunitx}

\usepackage{floatrow}

\title{Image-based Regularization for Action Smoothness in Autonomous Miniature Racing Car with Deep Reinforcement Learning}

\author[1]{Hoang-Giang Cao$^*$}
\author[1]{I Lee$^*$}
\author[1]{Bo-Jiun Hsu}
\author[1]{Zheng-Yi Lee}
\author[1]{Yu-Wei Shih}
\author[2]{\\Hsueh-Cheng Wang}  
\author[1,3]{I-Chen Wu$^\dagger$}
\affil[1]{Department of Computer Science, National Yang Ming Chiao Tung University, Taiwan}
\affil[2]{Department of Electrical and Computer Engineering, National Yang Ming Chiao Tung University, Taiwan}
\affil[3]{Research Center for IT Innovation, Academia Sinica, Taiwan}
\begin{document}
\maketitle

\def\thefootnote{$*$}\footnotetext{Equal contribution.}\def\thefootnote{\arabic{footnote}}
\def\thefootnote{$\dagger$}\footnotetext{Correspondence.}\def\thefootnote{\arabic{footnote}}

\thispagestyle{empty}
\pagestyle{empty}


\begin{abstract}
Deep reinforcement learning has achieved significant results in low-level controlling tasks.
However, for some applications like autonomous driving and drone flying, it is difficult to control behavior stably since the agent may suddenly change its actions which often lowers the controlling system's efficiency, induces excessive mechanical wear, and causes uncontrollable, dangerous behavior to the vehicle. 
Recently, a method called conditioning for action policy smoothness (CAPS) was proposed to solve the problem of jerkiness in low-dimensional features for applications such as quadrotor drones.
To cope with high-dimensional features, this paper proposes image-based regularization for action smoothness (I-RAS) for solving jerky control in autonomous miniature car racing. 
We also introduce a control based on impact ratio, an adaptive regularization weight to control the smoothness constraint, called IR control.
In the experiment, an agent with I-RAS and IR control significantly improves the success rate from 59\% to 95\%.
In the real-world-track experiment, the agent also outperforms other methods, namely reducing the average finish lap time, while also improving the completion rate even without real world training.
This is also justified by an agent based on I-RAS winning the 2022 AWS DeepRacer Final Championship Cup.

\end{abstract}


\section{Introduction}

In recent years, deep reinforcement learning (DRL) has achieved many milestones, particularly for games such as AlphaGo \cite{silver2016alphago}, AlphaStar \cite{Vinyals2019Grandmaster}, OpenAI Five \cite{openai2019dota}. 
DRL has also been applied to many real world applications.
Shixiang Gu {\em et al.} applied DRL for robotic manipulation with asynchronous off-policy updates \cite{Gu2017off}.
The work in \cite{PI2020104222} used low-level control through reinforcement learning directly to motors output. 
For autonomous driving, Amazon Web Services (AWS) provided DeepRacer, an autonomous racing experimentation platform for sim-to-real reinforcement learning \cite{Bharathan2019DeepRacer}.

However, a critical problem in control policy in real world applications is the jerky behavior, when training the agent in a complex dynamic environment \cite{Madhmood2018Benchmarking, Molchanov2019Transfer}.
Especially, in autonomous driving or quadrotor drone flying, jerky control causes many serious problems, such as uncontrollable movement and power consumption, which reduce the service life of the autonomous vehicle.
Prior works addressed the issue of smoothness policy by using reward engineering \cite{Hwangbo2017Control, Koch2019Flight, carlucho201871}.
Their approach designed a reward function for a specific task.
In autonomous driving, for example, the agent will be penalized if the current action is too different from the previous action, or the selected speed is too slow.
Reward engineering is based on prior human knowledge about the tasks. 

Recent research used DRL algorithm to solve this problem, trying to maximize total episode reward, and also smoothing control or action oscillation \cite{Yu2021TAAC, Chen2021NestedSoftAC}.
Siddharth Mysore {\em et al.} proposed conditioning for action policy smoothness (CAPS) for solving jerky actions by adding regularization terms\cite{siddharth2021caps}.
CAPS was originally used to smooth the control of quadrotor drones with some low-dimensional features.
So far, CAPS has not been applied to solving the jerky control problem in autonomous driving, particularly for image-based control. 
In the autonomous racing system with image-based control, the environment is more complex and dynamic due to environmental conditions and physical constraints.
Besides, dealing with high-dimensional input in reinforcement learning is very challenging and difficult due to sample inefficiency \cite{cai2021visionbased}.
To cope with high-dimensional features, this paper proposes image-based regularization for action smoothness (I-RAS) to solve jerky control in autonomous miniature car racing.

The main contributions of this paper are summarized as follows:
\begin{inparaenum}[1)]
    \item We propose I-RAS for solving jerky control in autonomous miniature car racing with image-based control.
    \item We introduce a control based on impact ratio, an adaptive hyperparameter to dynamically control the impact of smoothness constraint in I-RAS.
    \item The agent with I-RAS and IR control outperforms other methods, namely reducing the average finish lap time, while also improving the success rate, even in a real environment without real world training.
\end{inparaenum}

\section{Backgrounds}
\subsection{Smoothness Control}
For the issue of smoothness as described in the previous section, some prior works addressed it by using reward engineering \cite{Madhmood2018Benchmarking}.
The works in \cite{Hwangbo2017Control} and \cite{Molchanov2019Transfer} defined the reinforcement learning cost function based on the internal states of the quadrotor, like position, angular velocities, linear velocities, etc.
However, to design a reward function for a specific task, we may need to access the information from the environment, which may not be readily accessible.

Recent researches focus on the regularization penalties as a condition to achieve an action-smoothness policy.
Liu et al \cite{liu2021regularization} studied the regularization techniques to improve the performance of control policy on continuous control tasks.
Yu {\em et al.} presented temporally abstract actor-critic TAAC, an off-policy reinforcement learning incorporated with closed-loop action repetition a kind of temporal abstraction \cite{Yu2021TAAC}.
TAAC learned to select different actions only at critical states as possible.
A work in \cite{Chen2021NestedSoftAC} proposed Nested Soft Actor-Critic (NSAC), a DRL algorithm that helps to reduce oscillation behavior in autonomous driving.
This approach is only designed for discrete policies, while our task is for continuous policy.
Dmytro {\em et al.} \cite{Dmytro2019Autoregressive} researched the smooth exploration trajectories in continuous control tasks with low-dimensional input.

Siddharth Mysore {\em et al.} \cite{siddharth2021caps} introduced conditioning for action policy smoothness (CAPS) and got significant improvements in controller smoothness and power consumption. 
The authors proposed spatial and temporal smoothing regularization terms to optimize policy directly.
Originally, CAPS was applied to drones and used internal states as the input, which is hard to be used for applications that require visual input like autonomous driving.
A work in \cite{Axe2021latent} applied temporal smoothness (of CAPS) to their work in autonomous racing, without using spatial smoothness. 
They reported that the agent achieved smoother steering, but did not improve the performance compared to the one trained without regularization.
This paper in contrast proposes an extensive method so as to smooth and stabilize the behavior while improving the performance in terms of success rate and finish lap time, as described in the rest of this paper.

\subsection{Sim-to-real Transfer}
Training DRL usually requires a large amount of data for training and collecting data in the real world is extremely costly and time-consuming.
In addition, to acquire an optimal policy or near-optimal policy, the agent needs to perform random actions to explore, which can be unstable and dangerous during real world data collection for a task like autonomous driving.
Sim-to-real provides an efficient approach for training DRL when we train agents in the simulation and migrate them to the real world.
However, the intrinsic gap between simulation and the real world often causes the agent trained in simulation to perform poorly in the real world \cite{hofer2020perspectives}.
Prior works mainly focused on domain randomization and domain adaptation methods.

The most widely used sim-to-real transfer method is domain randomization.
Domain randomization methods try to enlarge the distribution of domain-randomized simulation images to cover real world data \cite{zhao2020sim}.
The work in \cite{chu2020sim} proposed a student-teacher method for sim-to-real transfer in DeepRacer.
Domain randomization usually requires careful task-specific hyper-parameter tuning, and the cost grows exponentially with the number of parameters \cite{hofer2020perspectives}.
A recent method, named random convolution (RandConv)\cite{xu2022randomconv}, where the weights are randomly sampled from a Gaussian distribution at each iteration.

\section{Methods}
\subsection{Conditioning for Action Policy Smoothness: CAPS}
\label{sec:caps}

\begin{figure}[t]
        \includegraphics[width=0.8\columnwidth]{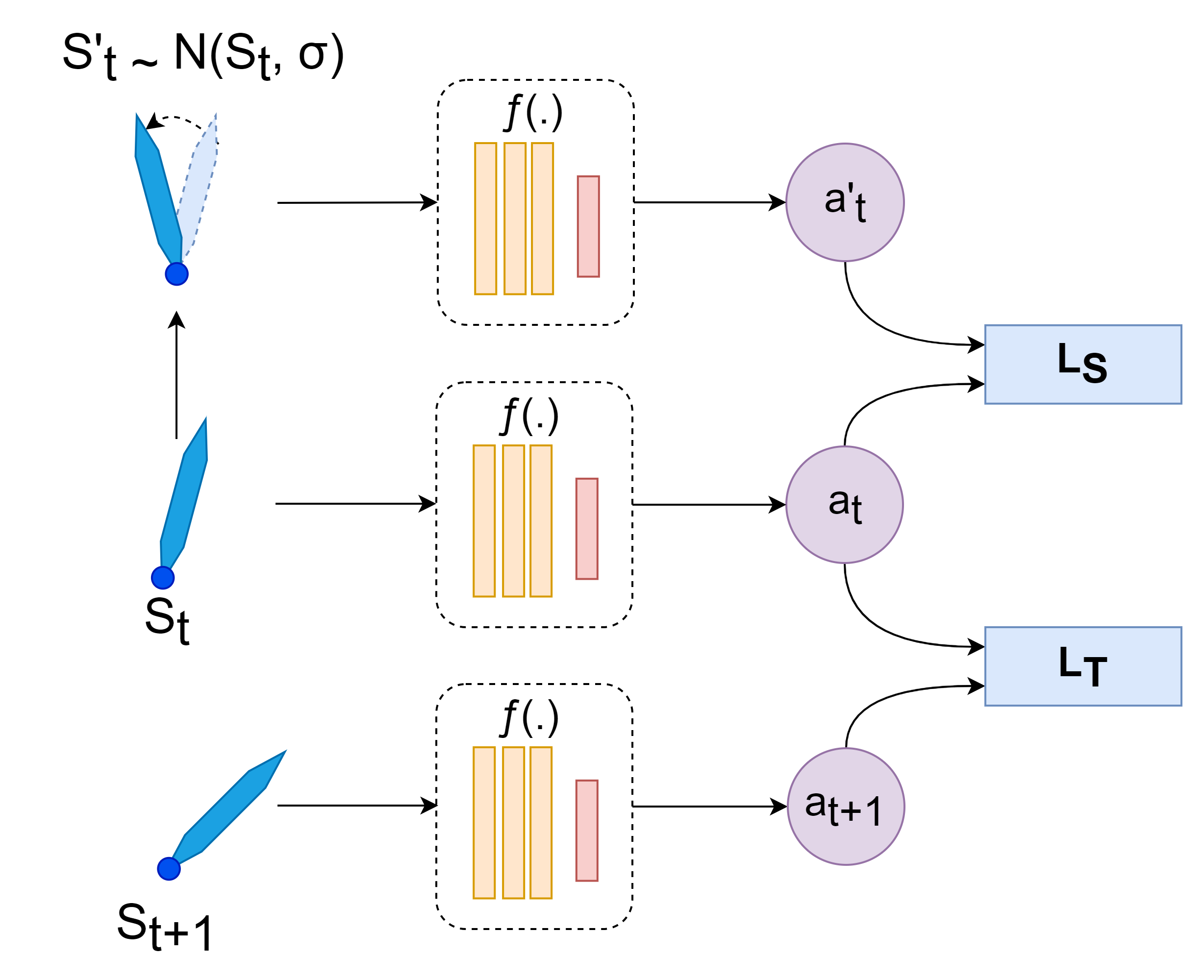}
    \caption{Overview of the regularization terms of CAPS with low-dimensional input, where the share $f(.)$ maps from state to action.
    The temporal smoothness $L_{T}$ penalizes divergent actions between consecutive states.
    The spatial smoothness $L_{S}$ encourages consistent actions on similar states $S'_t$, where $S'_t$ is drawn from a normal distribution around the current state $S_t$ with standard deviation $\sigma$.
    }
    \label{fig:caps_smoohness}
\end{figure}

Mysore {\em et al.} \cite{siddharth2021caps} introduced conditioning for action policy smoothness (CAPS) and got significant improvements in controller smoothness and power consumption on a quadrotors drone.
To condition policies for smooth control, the authors proposed two regularization terms:
\begin{inparaenum}[1)]
    \item temporal smoothness term, and 
    \item spatial smoothness term.
\end{inparaenum}


The policy $\pi$ is a mapping function of states $s$ to actions $a = \pi(s)$.
The objective function of CAPS, $J_{\pi}^{\scalebox{0.5}{CAPS}} $, contains three components: a reinforcement learning objective function $J_{\pi}$; temporal smoothness regularization term $L_{T}$; and spatial smoothness regularization term $L_{S}$.
The regularization weights $\lambda_{T}$ and $\lambda_{S}$ are used to balance the impact of two regularization terms $L_{T}$ and $L_{S}$, respectively.

\begin{equation}
J_{\pi}^{\scalebox{0.5}{CAPS}} = J_{\pi} - \lambda_{T}L_{T} - \lambda_{S}L_{S}
\label{equation:objective_functino_caps}
\end{equation}
\begin{equation}
L_{T} = D_{T}(\pi(s_{t}),\pi(s_{t+1}))
\end{equation}
\begin{equation}
L_{S} = D_{S}(\pi(s_{t}),\pi({s'}_{t})) \quad where \quad s'_t\sim \Phi(s_{t}).
\end{equation}
Both $D_T$ and $D_S$ are calculated based on the Euclidean distance of two vectors.
The temporal smoothness term $L_{T}$ penalizes the $J_{\pi}^{\scalebox{0.5}{CAPS}}$ when the action of the next states $s_{t+1}$ are significantly different from the actions of the current states $s_{t}$.
The spatial smoothness term $L_{S}$ encourages the policy to take similar actions on the similar states $s'_t$, which are drawn from a distribution $\Phi$ around states $s_t$, namely \(\Phi(s) = N(s,\sigma)\) with standard deviation \(\sigma\)  around $s$. 
Originally, CAPS is applied to smooth the control of a quadrotor drone, with the input being the inertial measurement unit (IMU) and the electronic speed controller (ESC) sensors.
\autoref{fig:caps_smoohness} illustrates a design for temporal smoothness and spatial smoothness in CAPS.


\begin{figure*}
\centering
    \begin{subfigure}[t]{0.5\columnwidth}
        \includegraphics[width=\columnwidth]{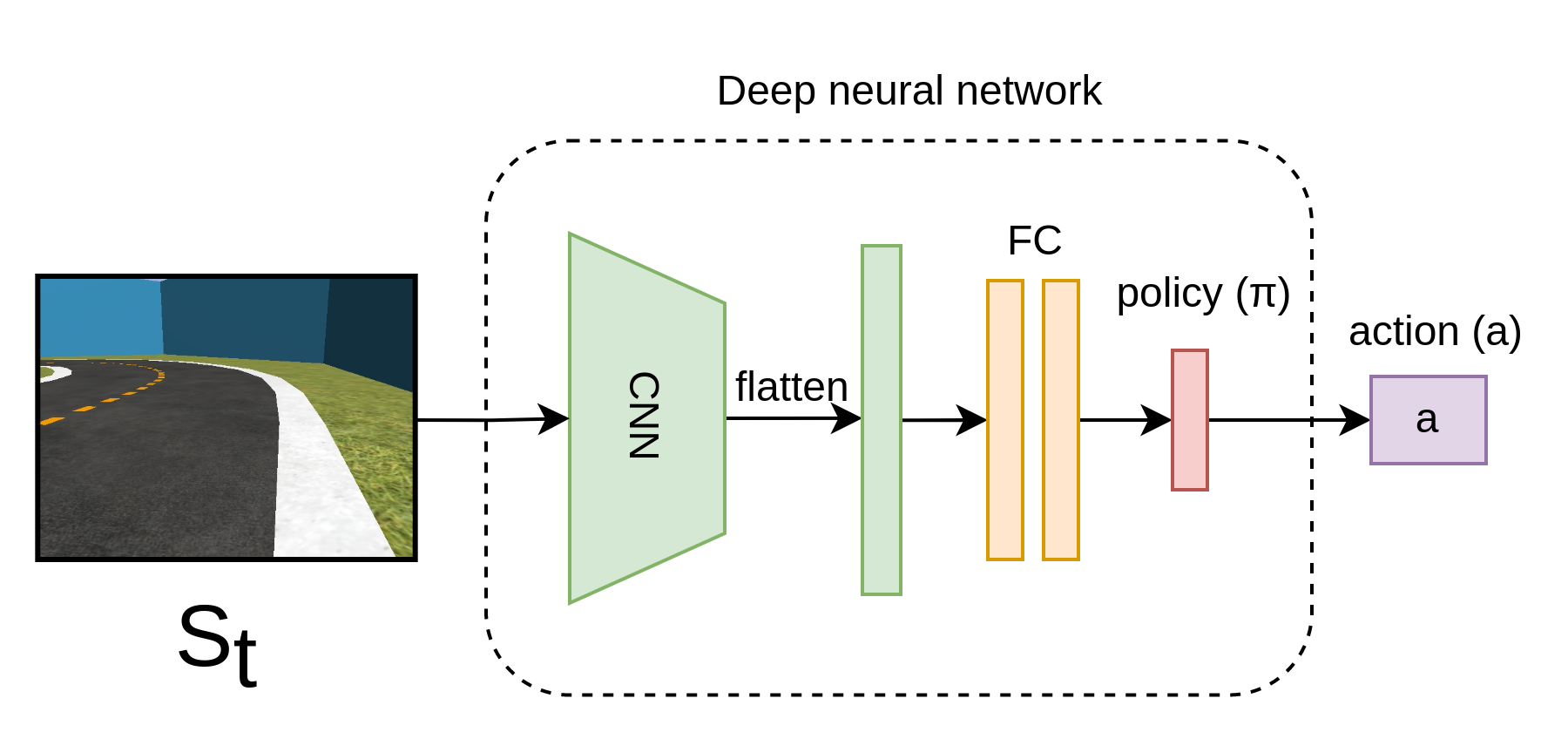}
    \caption{Network structure.}
    \label{fig:network_structure}
    \end{subfigure}
    \begin{subfigure}[t]{0.4\columnwidth}
        \includegraphics[width=\columnwidth]{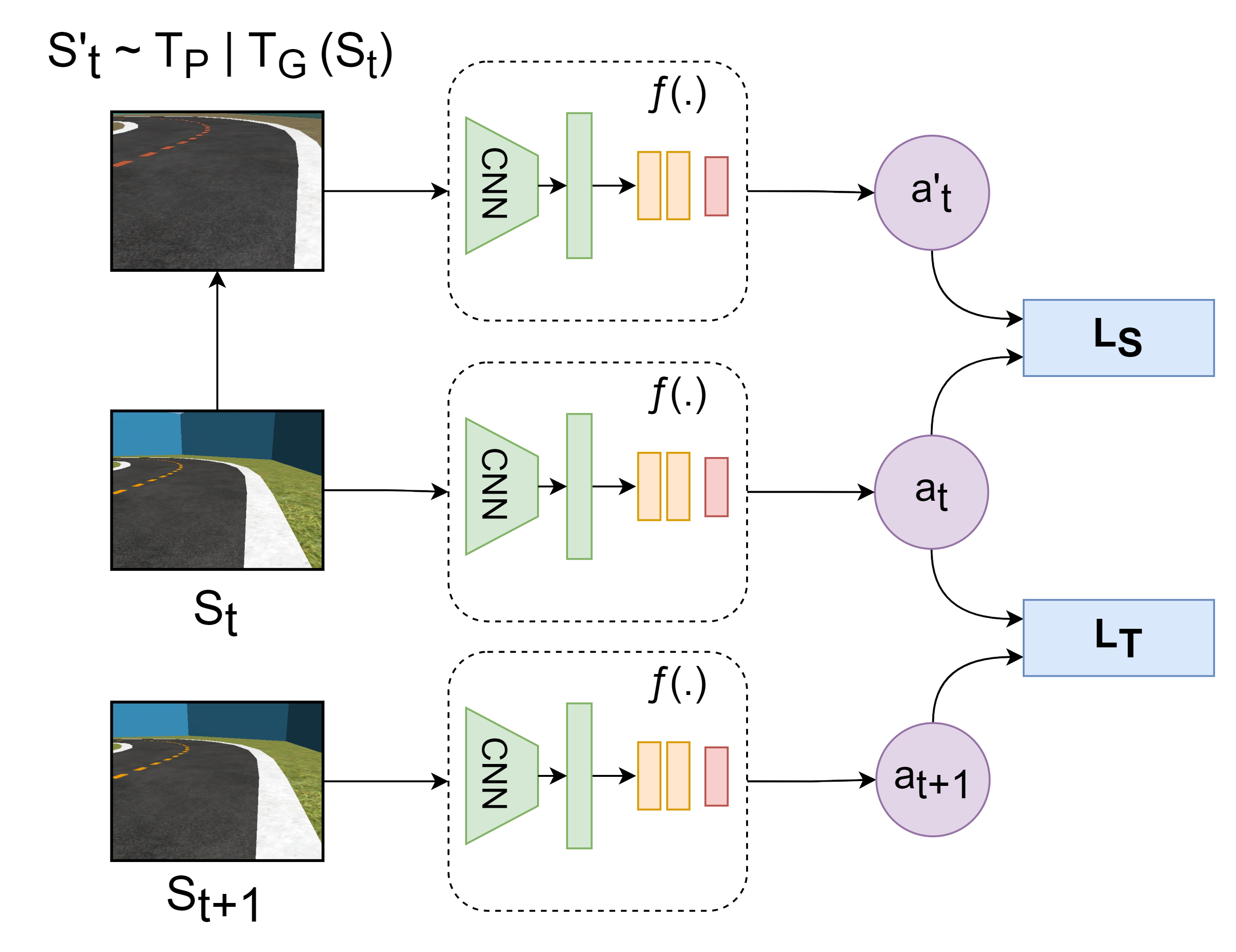}
        \caption{Overview of I-RAS.}
        \label{fig:IRAS_overview}
    \end{subfigure} 
\caption{(a) Policy network structure with three convolutions layers followed by three fully connected layers. The output is continuous action for normalized values of steering and speed.
(b) The overview of regularization terms of I-RAS with image-based input.
Temporal smoothness $L_{T}$ and spatial smoothness $L_{S}$ hold a similar meaning as in CAPS.
For image input, we generate a similar state $S'_t$ from the current state $S_t$ using either photometric transformation $T_P$ or geometric transformation $T_G$.}
\label{fig:i-ras-overview}
\end{figure*}

\begin{table*}[ht]
\begin{tabular}{|c|c|c|l|}
\hline
Normalized speed & Normalized reward & IR control value & Description                                                                                                                                    \\ \hline
1.0              & 1.0               & 1.0          & \begin{tabular}[c]{@{}l@{}}High speed, good action (high reward).\\ Maintain the action.\end{tabular}                        \\ \hline
1.0              & 0.1               & 0.316          & \begin{tabular}[c]{@{}l@{}}High speed, bad action (low reward).\\ Need to change the action; otherwise, out of the track.\end{tabular} \\ \hline
0.1              & 1.0               & 0.316          & \begin{tabular}[c]{@{}l@{}}Low speed, good action (high reward).\\ Need to change the action to make a turn.\end{tabular}         \\ \hline
0.1              & 0.1               & 0.1         & \begin{tabular}[c]{@{}l@{}}Low speed, bad action (low reward).\\ Definitely need to change the action.\end{tabular}                 \\ \hline
\end{tabular}
\caption{Illustrations of normalized speed and reward, and IR control value together with a description of their relation to the smoothness regularization constraint in the last column.
A high value of IR control means that the agent should maintain the action for smoothness control.
A low value of IR control means that the agent should change the action since the current action is inappropriate or need to make a turn.}
\label{tab:explain_impact_ratio}
\end{table*}


\subsection{Image-based Regularization for Action Smoothness in Autonomous Racing Car: I-RAS}
\label{sec:method_iras}
CAPS was originally applied to smooth the control of a drone with the internal states of the rotors, which does not directly fit the applications that require visual input like autonomous driving.
In order to generate the similar state $s'_t$ in spatial smoothness in CAPS, the authors \cite{siddharth2021caps} used Gaussian Noise to draw $s'_t$ from a normal distribution around state $s_t$ as described in Subsection \ref{sec:caps}.
Now consider the applications that need to use images as the input. 
A straightforward method is to draw from a normal distribution (Gaussian noise) on image pixels like the original CAPS. 
Actually, we can apply domain randomization to generate a similar state $s'_t$ in the spatial smoothness, instead of drawing from a normal distribution.

In this paper, we categorize domain randomization into two types: photometric transformation and geometric transformation.
The photometric transformation uses filters to change the intensity of the image.
In the autonomous racing car, the photometric transformation simulates the different conditions of the environment like light, darkness, shadows, flare, etc., or camera conditions like noise, blurred image, etc.
We use four photometric transformation methods $T_P$: random brightness, random contrast, salt and pepper, and Gaussian blur.
The geometric transformation is used to generate a possible next state that the agent might observe in the next step.
We use three geometric transformation methods $T_G$: rotation, shift, and scale.
We then generate a similar state  $s'_t$ by using either geometric transformation $T_G$ or photometric transformation $T_P$ from the current state $s_t$.

\begin{figure}[ht]
    \includegraphics[width=0.95\columnwidth]{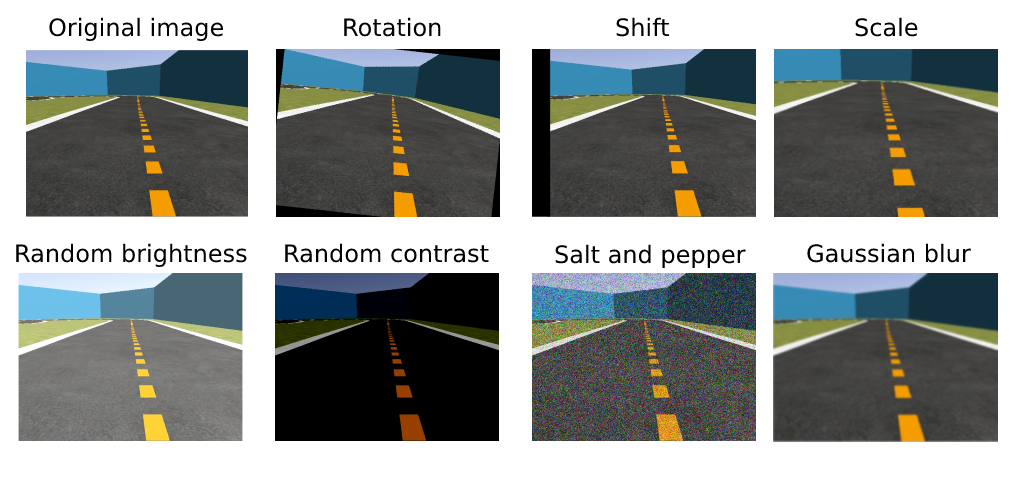}
\caption{Example images generated by geometry transformation (the upper-right three) and photometric transformation (the lower four) used in the spatial smoothness.}
\label{fig:domain_randomization}
\end{figure}

The spatial smoothness in I-RAS is defined as:
\begin{equation}
L_{S} = D_{S}(\pi(s_{t}),\pi({s'}_{t})) \quad where \quad s'_t\sim T_P|T_G (s_{t}).
\end{equation}



For temporal smoothness $L_T$, we also penalize the policy in the same way as CAPS, when the agent applies significantly different actions between two continuous steps.
\autoref{fig:network_structure} illustrates our network structure and \autoref{fig:IRAS_overview} shows the overview of I-RAS with image-based input. 
\autoref{fig:domain_randomization} shows some examples images that are generated by $T_G$ and $T_P$.

\subsection{Flexible Smoothness Action Constraint with IR Control}

\begin{figure}[!b]
    \includegraphics[width=0.6\columnwidth]{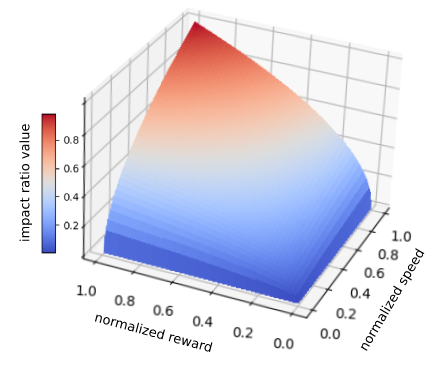}
\caption{Visualize IR control value with respect to normalized reward and speed. Red color indicates high value and blue color indicates low value.}
\label{fig:plot_impact_ratio}
\end{figure}

CAPS strictly made the policy excessively smooth which lost the capability to handle situations that require the agent to change the action.
Especially in autonomous racing, the car needs to quickly reduce the speed and change the steering action at a curve.
To solve this problem, we introduce the impact ratio control (IR control) $\lambda_{IR}$, an adaptive regularization weight that combines the normalized speed with the normalized reward of the current state:

\begin{equation}
\lambda_{IR} =\sqrt{ speed^{*}_t \times reward^{*}_t}
\end{equation}
The $ speed^{*}_t$ is the normalized speed in a range from [0,1] and the $reward^{*}_t$  is the normalized reward  in a range [0,1] of a state $s_t$.
A lower value of $\lambda_{IR}$ relaxes the smoothness constraint and allows the agent to substantially change the action.
A higher value of $\lambda_{IR}$ will maintain the current action for smoothness trajectory.
\autoref{tab:explain_impact_ratio} shows some examples of IR control value and explains the relation to the smoothness regularization constraint.
The loss function of I-RAS with IR control is defined as:
\begin{equation}
J_{\pi}^{\scalebox{0.5}{IR}} = J_{\pi} -  \lambda_{IR} (\lambda_{T}L_{T}  + \lambda_{S}L_{S})
\label{equation:objective_functino_iras_ir}
\end{equation}
\autoref{fig:plot_impact_ratio} visualizes how the IR control value is related to the normalized speed value and normalized reward value.

\begin{figure*}[ht]
    \centering
        \includegraphics[width=0.95\textwidth]{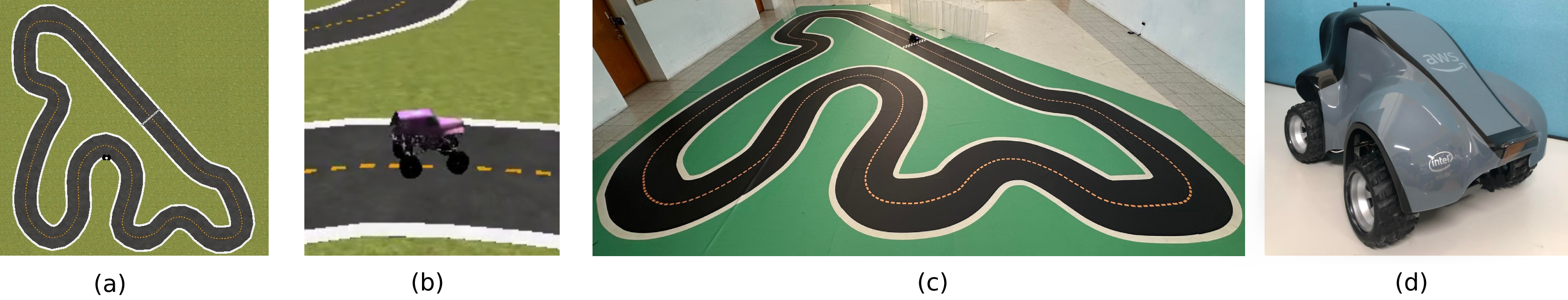}
    \caption{Environment setup in simulation and real world of AWS DeepRacer. (a) and (b) are the racing track and car in simulation.
    (c) and (d) are the racing track and the car in our own real-world environment, where 
    there is an intensely difficult track (35.87 meters) with many devious curves.}
    \label{fig:aws_env_setup}
\end{figure*}

\subsection{Sim to Real Transfer: RandConv}
\label{sec:randconv}
In this paper, we apply random convolution (RandConv) \cite{xu2022randomconv} as the sim-to-real transfer method.
In contrast to other fixed weights of filters in most of the other domain randomization methods, RandConv is a data augmentation technique using multi-scale random convolutions to generate images with random textures while maintaining global shapes.
The weights of filters in RandConv are randomly sampled from a Gaussian distribution at each iteration.
This is a promising approach that can replace other hand-picked domain randomization methods.
Since this paper focuses on action smoothness, this paper uses it as a standard domain randomization method in all real-world cases for simplicity of analysis. 

\section{Experiment Setup}
\label{sec:experiments_setup}
\subsection{Environments}

Our experiments are done on DeepRacer \cite{Bharathan2019DeepRacer}, an autonomous car racing platform developed by AWS, as shown in \autoref{fig:aws_env_setup}.
The AWS DeepRacer uses a 1/18th scale race car compared to a real-sized race car, which provided a more affordable way to conduct research in autonomous racing.
The agent receives the observation from a camera placed on the top of the miniature car racing.
The observation is an RGB image with a resolution of 120x160, and each pixel color ranges from 0 to 255.
For the action space, it is free to choose between discrete actions and continuous actions.
We use the continuous space where the output includes the normalized steering angle in a range from [-1, 1], covering from -30 to 30 degrees, and normalized speed with a range from [-1, 1], covering from 1m/s to 4m/s based on the configuration of AWS DeepRacer. 

For the baseline, we implement a version of
Soft Actor-Critic (SAC) \cite{Haarnoja2018SAC} with the following configurations: three workers are used; gamma is 0.98; initial alpha is 0.3; batch size is 1024; learning rate is 0.0003; global buffer size is 10000; local buffer size is 2000.
Following the configuration in \cite{siddharth2021caps}, the regularization weights of temporal smoothness and spatial smoothness in Equation (\ref{equation:objective_functino_caps}) are set to $\lambda_T = 1$ and $\lambda_S = 5$ respectively.

\begin{table*}[!ht]
\centering
\begin{tabular}{lcccc}
\hline \hline

Agent & Success rate (\%)${\uparrow}$ & Finish lap time (s) ${\downarrow}$ & Avg speed ${\uparrow}$  & Steering $S_m$ $ {\downarrow}$  \\ 
\hline

Vanilla SAC &       59\% &       18.12 ± 0.63                                   & 1.37                 & 0.070                                                                      \\ 
                                 
I-RAS       & 87\% &        13.34 ± 0.35                                    & 2.12                        & 0.059                                                                   \\ 
I-RAS + IR control & \textbf{95\%} &   \textbf{ 13.10 ± 0.34}                                                       & \textbf{2.23}     & \textbf{0.058}                                                     \\ \hline \hline
\end{tabular}
\caption{
The performances on finish lap time, average speed, and steering smoothness value among different agents.
The I-RAS agent with IR control achieved a 95\% success rate with the fastest finish lap time while maintaining smoothness value.
}
\label{tab:full_perfomance_sim}
\end{table*} 

\subsection{Evaluation}
To evaluate the effectiveness of the smoothness method, we use a metric $S_m$, called a smoothness value proposed by \cite{siddharth2021caps}, a method based on the Fast Fourier Transform (FFT) frequency spectrum defined as follows.
\begin{equation}
\label{equation:smoothness_function}
    S_{m} = \frac{2}{nf_{s}}\sum_{i=1}^{n}M_{i}f_{i},
\end{equation}
where $M_i$ denotes the amplitude of the frequency component $f_i$, and $f_s$ the sampling rate, set to $f_s=30$ in this paper.
In general, the lower the value, the smoother the action, since the low-frequency control is weighed less.

To evaluate the performance of car racing, we use the following metrics:
\begin{inparaenum}[(a)]
    \item Action smoothness value: the smoothness value of the action of the agent.
    \item Average finishing lap time: the average time to finish a run (in seconds).
    \item Success rate: the percentage of completed runs against all runs.
\end{inparaenum}
A run is terminated after the car moves in the wrong direction on the track, all wheels are out of the track, or when it successfully completes the track.

\section{Experiment Results}
\label{sec:experiments}

In the experiment, we first evaluate the control smoothness ability of I-RAS in the simulation in Section \ref{sec:smoothness_with_iras}.
We then test I-RAS in the real environment and report the performance of our agent at the 2022 AWS DeepRacer Final Championship Cup (Section \ref{sec:real_experiment}).
Finally, in Section \ref{sec:ablation_study}, we conduct extensive experiments to analyze the impact of I-RAS components.

\subsection{Smoothness with I-RAS}
\label{sec:smoothness_with_iras}
In this experiment, we investigate the policy smoothness capability of I-RAS in the simulation.
We implement a vanilla SAC agent as a baseline, and two SAC agents trained with I-RAS (called I-RAS agents in the rest of this section) and with and without IR control respectively.
For each agent, we pick the model trained up to 50,000 iterations and then try 100 runs using the model.
\autoref{tab:full_perfomance_sim} shows the performance results for three agents including success rate, average finish lap time, average speed, and steering smoothness value.
From the results, we see that both I-RAS agents outperform the vanilla SAC by a large margin in terms of both success rate and finish lap time.
Particularly, the I-RAS agent with IR control achieves the highest speed and finishes a lap in 13.10 seconds on average, while maintaining the highest success rate and keeping the lowest steering smoothness value.

For the two I-RAS agents, we observe the following phenomenon.
The I-RAS agent without IR control tries to maintain a strong regularization between two contiguous states, thus it does not reduce the speed and steering significantly at sharp curves so as to lose control to make a turn. 
In contrast, the I-RAS agent with IR control is able to make a quicker change in speed and steering to make a turn.
As a result, the I-RAS agent with IR control greatly improves the success rate from 87\% to 95\%, with slightly better finish lap time compared to the one without the control.

\begin{figure*}[ht]
\centering
    \begin{subfigure}[t]{0.32\columnwidth}
        \includegraphics[width=\columnwidth]{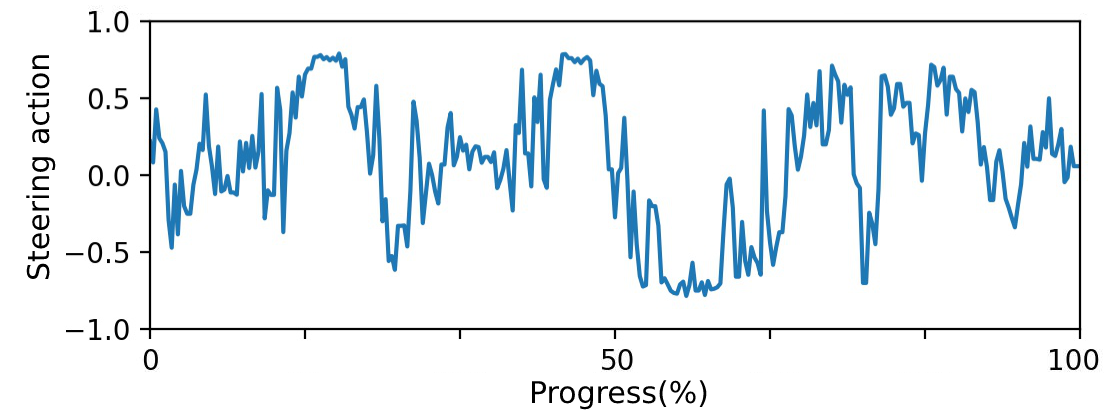}
        \caption{Vanilla SAC.}
    \label{fig:vanila_sac_steering}
    \end{subfigure}
    \begin{subfigure}[t]{0.32\columnwidth}
        \includegraphics[width=\columnwidth]{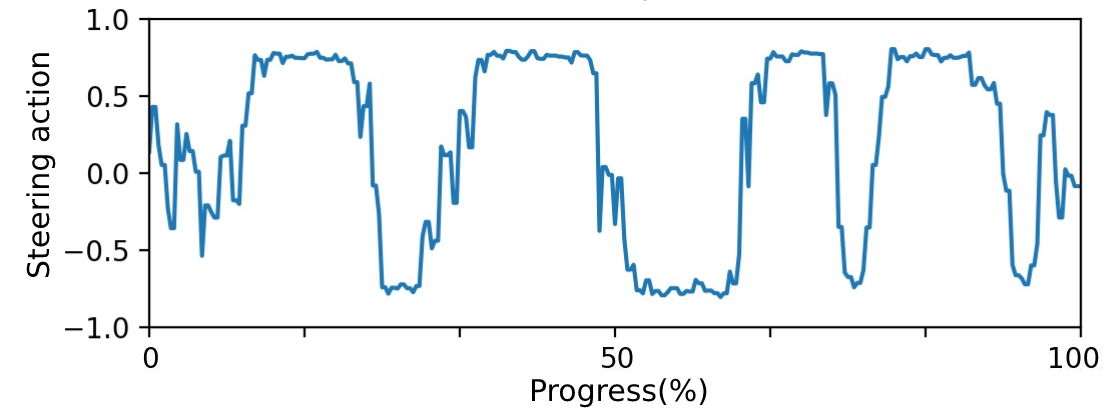}
        \caption{I-RAS.}
        \label{fig:ras_steering_1}
    \end{subfigure}
    \begin{subfigure}[t]{0.32\columnwidth}
        \includegraphics[width=\columnwidth]{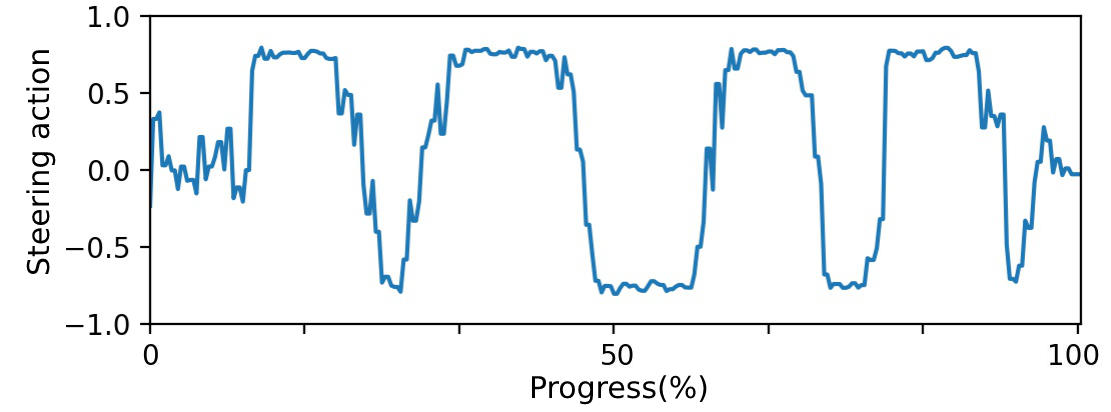}
        \caption{I-RAS with IR control.}
        \label{fig:ras_ir_steering}
    \end{subfigure}
    \begin{subfigure}[t]{0.32\columnwidth}
        \includegraphics[width=\columnwidth]{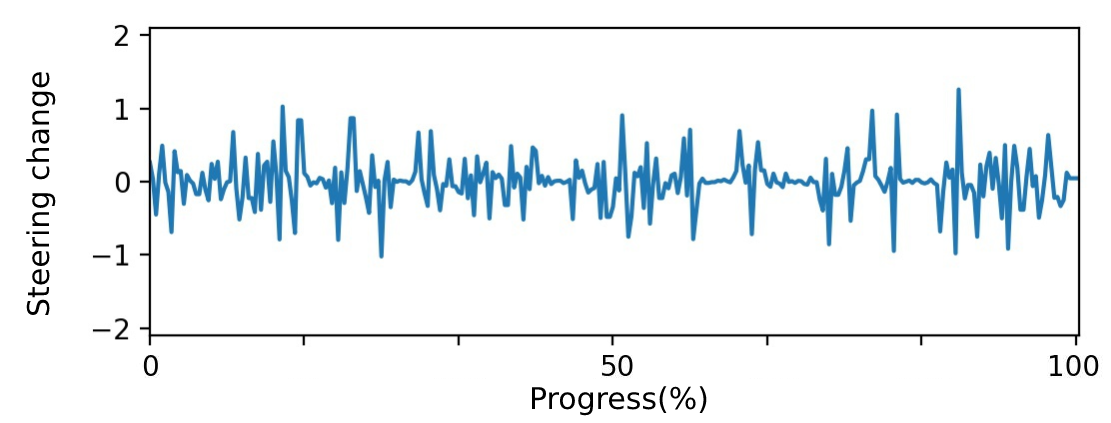}
        \caption{Vanilla SAC.}
    \label{fig:vanila_sac_steering_1}
    \end{subfigure}
    \begin{subfigure}[t]{0.32\columnwidth}
        \includegraphics[width=\columnwidth]{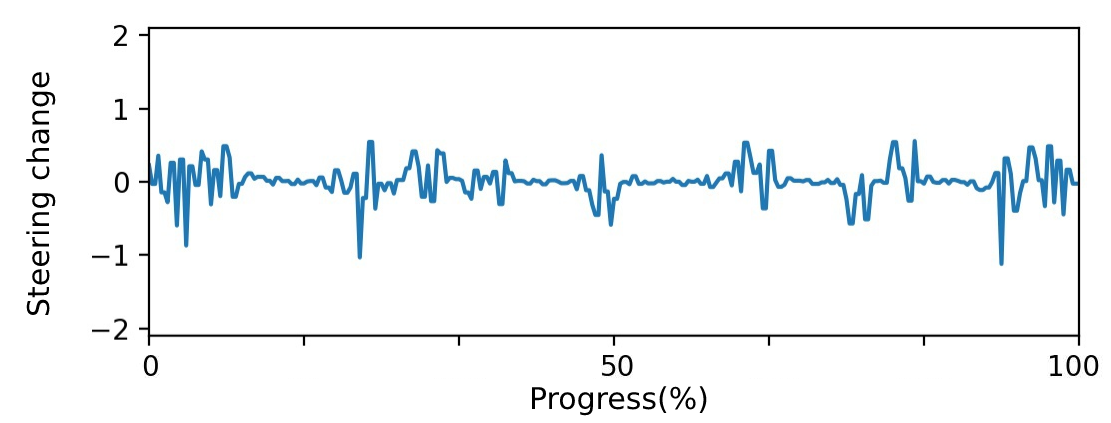}
        \caption{I-RAS.}
        \label{fig:ras_steering_2}
    \end{subfigure}
    \begin{subfigure}[t]{0.32\columnwidth}
        \includegraphics[width=\columnwidth]{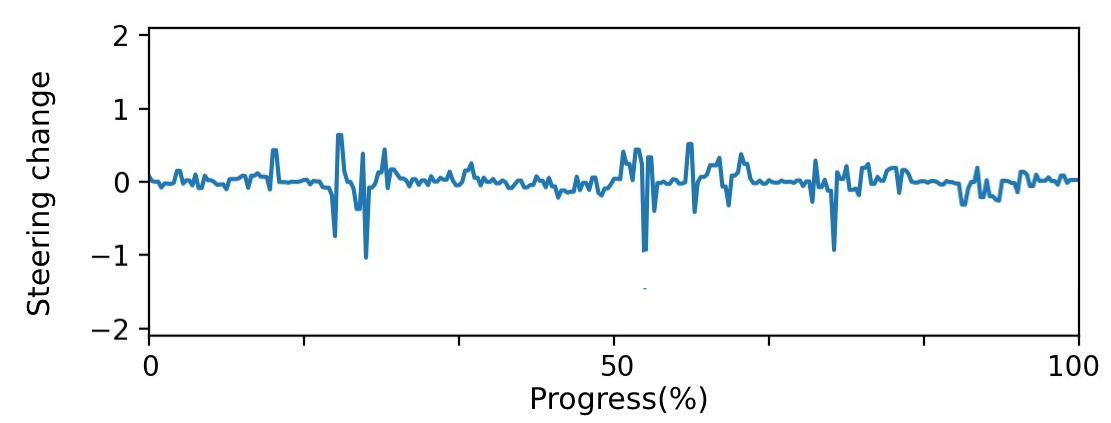}
        \caption{I-RAS with IR control.}
        \label{fig:ras_ir_steering_1}
    \end{subfigure}
    \begin{subfigure}[t]{0.32\columnwidth}
        \includegraphics[width=\columnwidth]{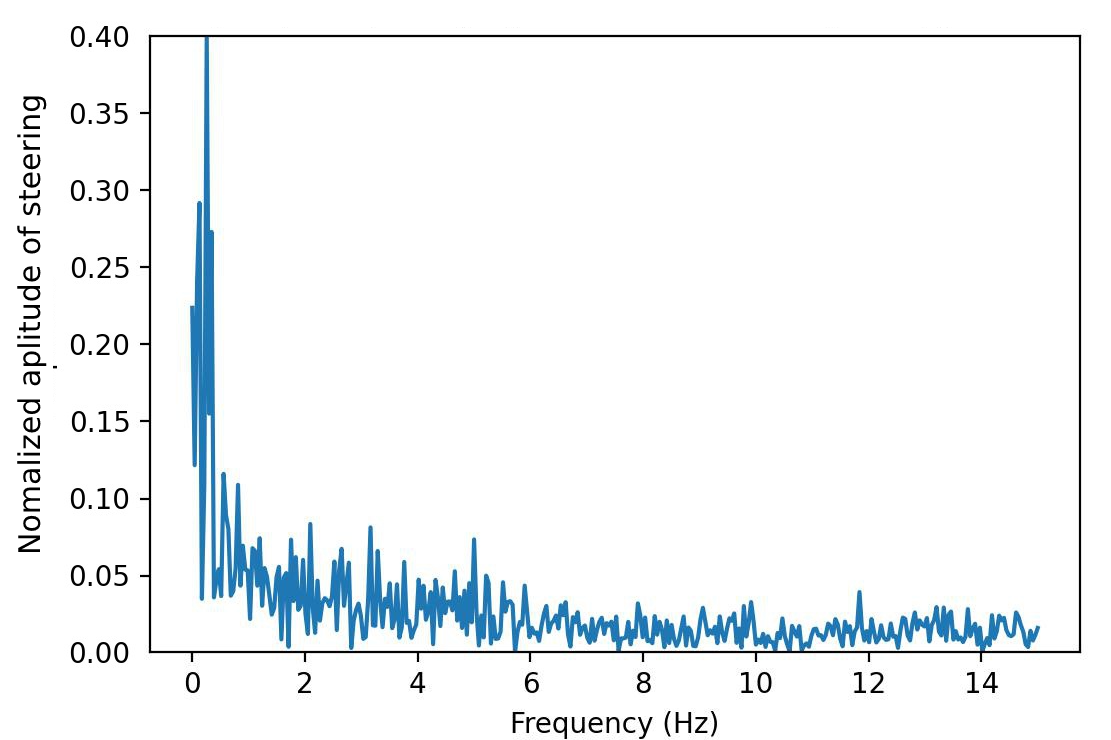}
        \caption{Vanilla SAC.}
    \label{fig:vanila_sac_steering_21}
    \end{subfigure}
    \begin{subfigure}[t]{0.32\columnwidth}
        \includegraphics[width=\columnwidth]{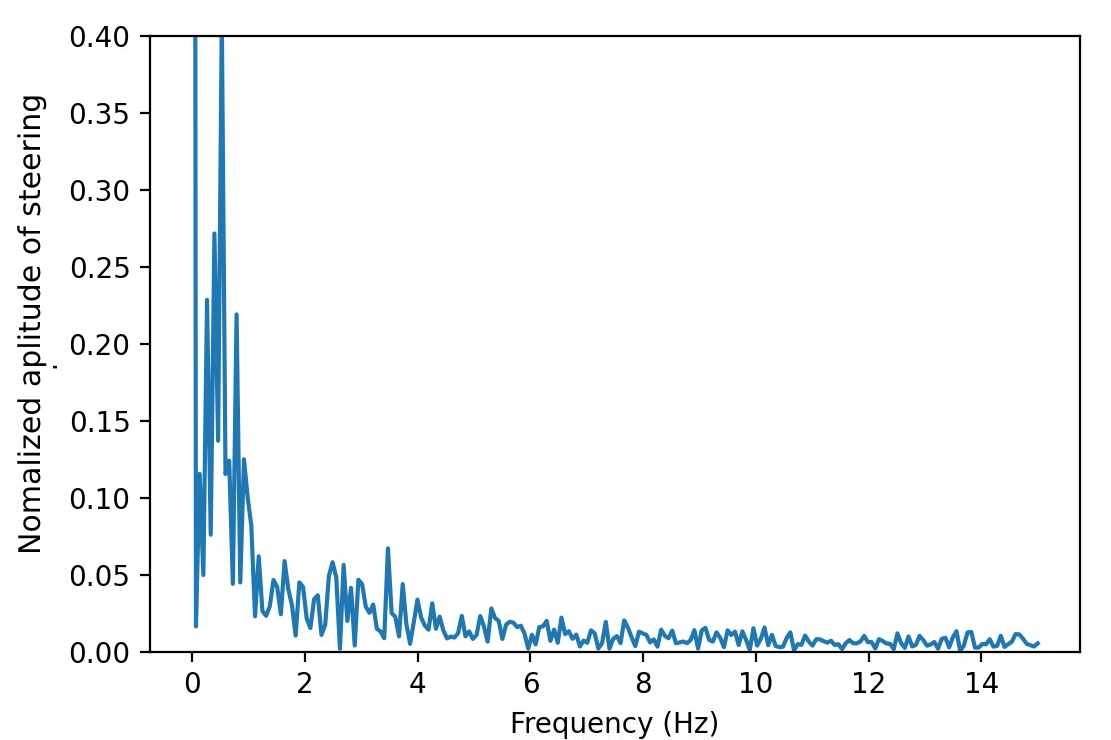}
        \caption{I-RAS.}
        \label{fig:ras_steering_32}
    \end{subfigure}
    \begin{subfigure}[t]{0.32\columnwidth}
        \includegraphics[width=\columnwidth]{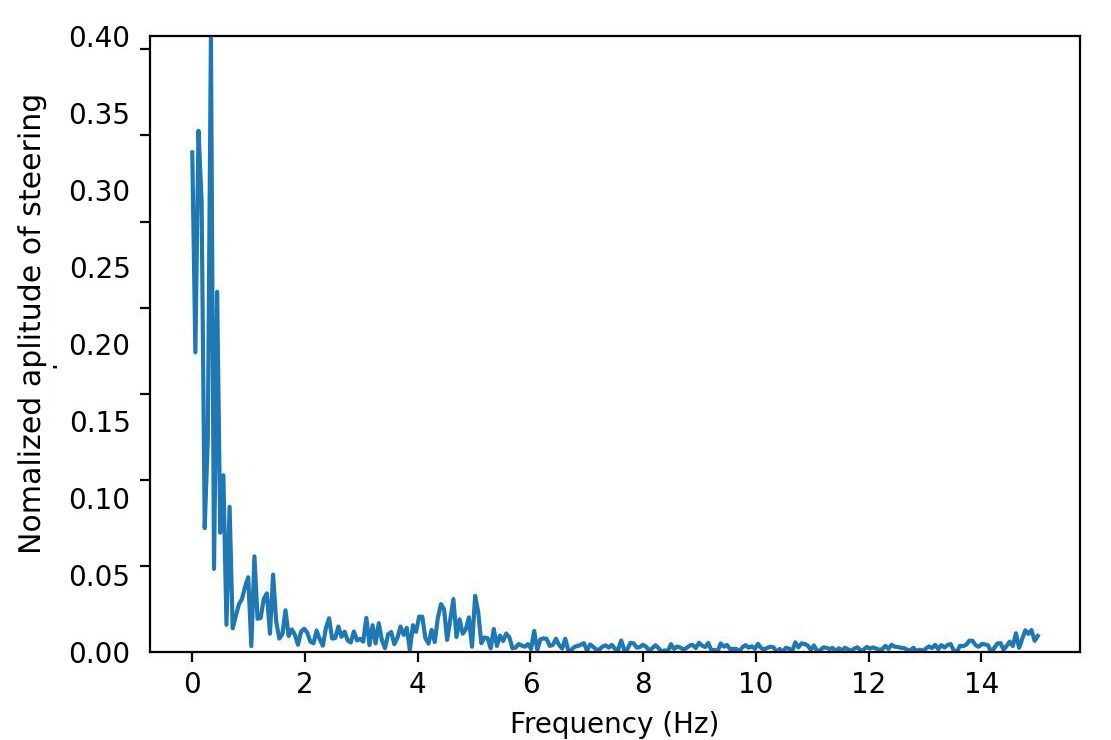}
        \caption{I-RAS with IR control.}
        \label{fig:ras_ir_steering_22}
    \end{subfigure}
        \begin{subfigure}[t]{0.32\columnwidth}
        \includegraphics[width=\columnwidth]{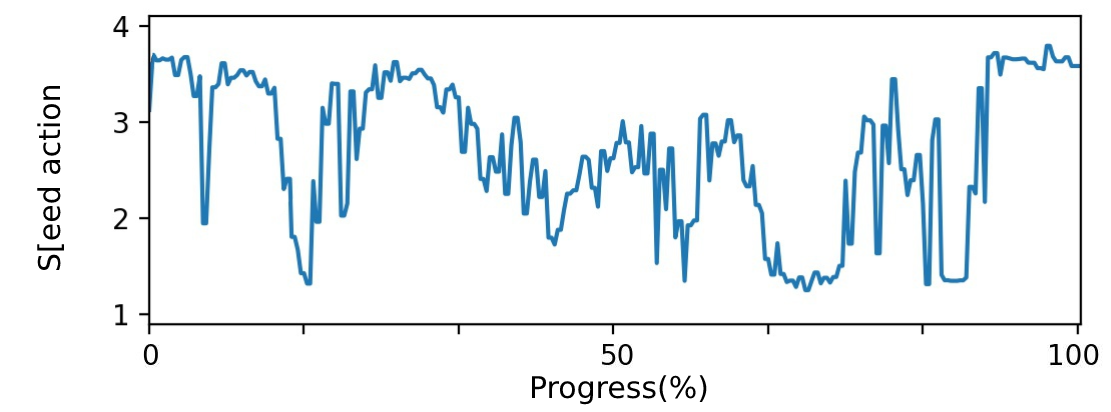}
        \caption{Vanilla SAC.}
    \label{fig:vanila_sac_steering_23}
    \end{subfigure}
    \begin{subfigure}[t]{0.32\columnwidth}
        \includegraphics[width=\columnwidth]{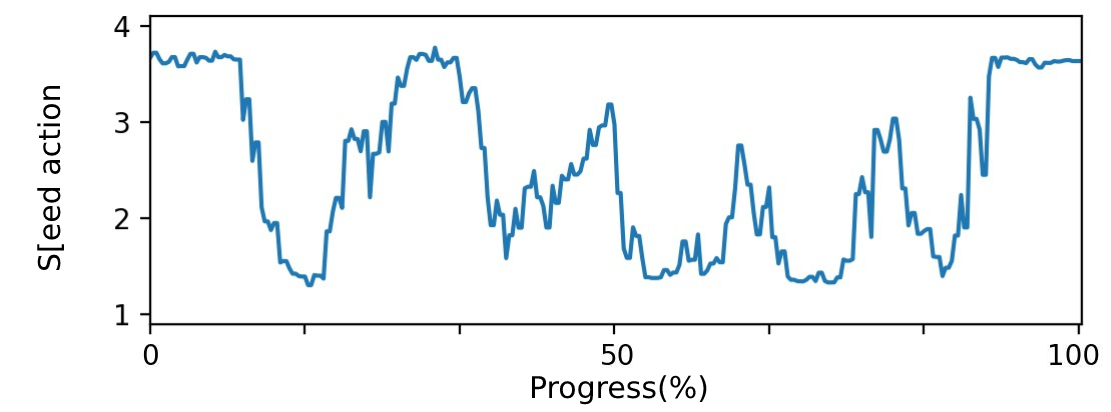}
        \caption{I-RAS.}
        \label{fig:ras_steering_31}
    \end{subfigure}
    \begin{subfigure}[t]{0.32\columnwidth}
        \includegraphics[width=\columnwidth]{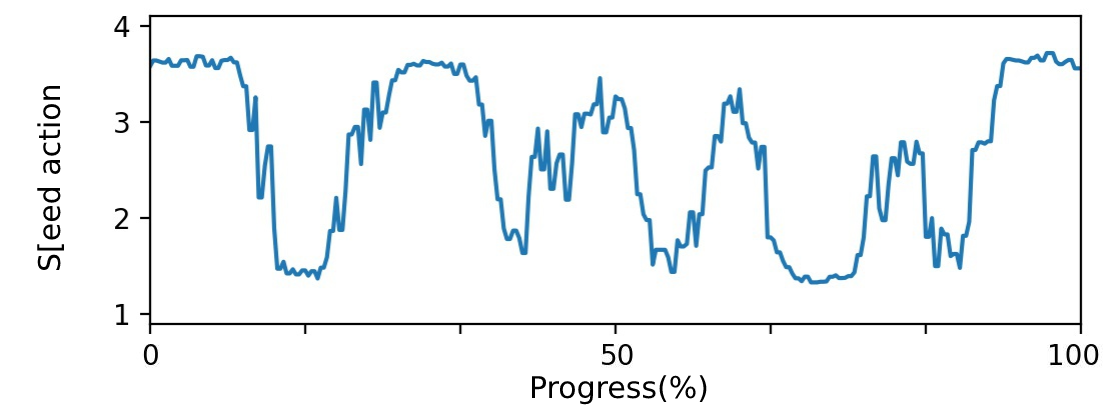}
        \caption{I-RAS with IR control.}
        \label{fig:ras_ir_steering_2}
    \end{subfigure}
\caption{Steering actions and corresponding steering changes,  FFTs of the three agents, vanilla SAC, I-RAS, and I-RAS with IR control (from left to right).
The progress is the normalized travel distances with respect to the track length of a run.
(a)-(c) are the steering actions through a completed track;
(d)-(f) the corresponding steering change;
 (g)-(i) the corresponding FFTs.
and (j)-(l) the speed action through a completed track.
With I-RAS, the agents significantly improve the smoothness of steering and speed compared to the vanilla SAC agent.
Therefore, in the corresponding steering FFTs, the agents with I-RAS show better stability with high-frequency components being reduced, compared to vanilla SAC.
}
\label{fig:full_compare_steering}
\end{figure*}

\autoref{fig:full_compare_steering} shows steering actions and their changes in detail during the progress of episodes. 
The agent with I-RAS has a better smoothness value of steering, which reduces from \num{0.070} in vanilla SAC to \num{0.059}.
By adding the adaptive regularization weight IR control, the agent still maintains a smooth trajectory while having the benefit to make a turn.
Therefore, the I-RAS agent with IR control not only maintains a smooth trajectory but also finishes a track in the fastest time.

\subsection{Real Experiments}
\label{sec:real_experiment}
We then evaluate our agent in the real work environment.
The agents were trained entirely in the simulation and tested in the real environment with zero real world training.
In the real test, each agent tries 15 attempts and calculates the average performance.
For the sim-to-real transfer, we simply apply RandConv as mentioned in Subsection \ref{sec:randconv}, a photometric transformation method for sim-to-real transfer.

\autoref{tab:real_experiment} shows, in the real environment, the success rate and finish lap time of the agents with the following models:
\begin{inparaenum}[(a)]
    \item Vanilla SAC.
    \item Vanilla PPO.
    \item I-RAS.
    \item I-RAS with IR control.
\end{inparaenum}
Note that we ignore the agent without RandConv since it cannot completely finish even for one run.
In general, the behaviors of I-RAS agents are more stable, finishing a run faster than the agents without I-RAS.
The two I-RAS agents have success rates at 60.00\% and 73.33\%, clearly outperforming the other two vanilla agents with SAC and PPO whose success rates are 46.66\% and 40.00\% respectively.
By adding the IR control, the finish lap time of the I-RAS agent is improved by 0.18 seconds in finish lap time.

\begin{table}
\begin{tabular}{lcc}
\hline \hline
\multicolumn{1}{c}{Agents} & \begin{tabular}[c]{@{}c@{}}Success\\ rate (\%)\end{tabular} & \begin{tabular}[c]{@{}c@{}}Finish lap\\ time (s)\end{tabular} \\ \hline
Vanilla SAC   &46.66\%   & 20.04        \\ 
Vanilla PPO  &40.00\%   & 22.03        \\ \hline
I-RAS &    60.00\%   & 19.54        \\ 
I-RAS + IR control & \textbf{73.33\%}    & \textbf{19.36}        \\ \hline \hline
\end{tabular}
\caption{Real experiments result with AWS DeepRacer.
The RandConv helps to transfer the model from simulation to the real environment.
The I-RAS agents with smoothness control perform better than other agents in terms of success rate and finish lap time.
}
\label{tab:real_experiment}
\end{table} 

\textbf{AWS DeepRacer Challenge.}
We then incorporated our proposed method into our agent and competed with others in the event of 2022 AWS DeepRacer Championship, where more than 150,000 developers participated according to the official reporters\footnote{AWS DeepRacer League \url{https://aws.amazon.com/deepracer/league/}}.
In the AWS DeepRacer Final Championship Cup 2022, our agent with I-RAS performed even better in the stadium in Las Vegas, namely finished a track much faster, reducing the lap finish time from about 19.360 to 13.756 seconds which is also the record in the whole competition. 
Consequently, our agent ILI-NYCU-CGI won the championship in this event.
From this event, interestingly, we also observed that differences in environmental conditions affect the performances significantly.
For example, the condition of friction at the stadium of the AWS Final Championship was much better than that in our own environment, so our agent can run even much faster due to the friction. 
Other factors may also include the conditions of environments, such as lights, the fences of tracks, and even the stability of miniature cars, but these factors are not in the research scope of this paper. 
A video clip is included in the supplementary to demonstrate (a) the performances of agents with and without I-RAS, and (b) the run of our I-RAS agent, the fastest finishing with 13.756 seconds in the AWS DeepRacer competition.

\subsection{Ablation study}
\label{sec:ablation_study}
\subsubsection{Temporal Smoothness and Spatial Smoothness}
\label{sec:ablation_study_iras_component}
In this ablation experiment, we study the impact of I-RAS components on the objective function in Equation (\ref{equation:objective_functino_iras_ir}) by ablating either temporal or spatial smoothness.
In addition to the agents with vanilla SAC and I-RAS (the same as the ones in Table \ref{tab:full_perfomance_sim}), we consider two more agents with the following models: 
\begin{inparaenum}[(a)]
    \item The agent with temporal smoothness term only (denoted as Temporal).
    \item The agent with spatial smoothness term only (denoted as Spatial).
\end{inparaenum}
The weights for both smoothness terms, $\lambda_T$ and $\lambda_S$, are set to 1.  

\autoref{tab:ablation_i_ras} lists the success rate, finish lap time, and steering smoothness of these agents, and shows that spatial smoothness improves the performances more significantly than temporal smoothness. 
In terms of finish lap time, both Spatial and I-RAS
require nearly the same time around 13 seconds, which are much faster than vanilla and Temporal around 18 seconds. 
However, with temporal smoothness added, it is negligible and even negligibly slower, namely slower by 0.04 seconds from Temporal to vanilla, and 0.08 seconds from I-RAS to Spatial. 
In terms of success rate, both Spatial and I-RAS (with spatial smoothness added) are also much better than vanilla and Temporal, and both I-RAS and Temporal are also better than Spatial and vanilla respectively, but not much. 
In terms of steering smoothness value, adding temporal smoothness does not show much difference in the results either, while adding spatial smoothness shows a significant reduction of the steering smoothness value.
This result concludes from experiments that spatial smoothness has a higher impact on the performance of the agent than temporal smoothness.


This is in contrast with the low-dimensional inputs experiment in CAPS, where the result of CAPS shows that temporal smoothness is more important than spatial smoothness.
In \cite{siddharth2021caps}, the authors used CAPS to control a quadrotor drone where the input is the low-dimensional input while we used image-based input which is high-dimensional input in autonomous racing.
In high-dimensional input, particularly in a stochastic environment like autonomous racing, the distribution of data is much more diverse than that in low-dimensional input.
We have the following conjectures. 
In low-dimensional input in CAPS, it is sufficient to obtain a smooth trajectory by learning temporal smoothness only from the observed next state.
However, in high-dimensional input, it is required to obtain a smooth trajectory by learning from possible next states generated in spatial smoothness.


\begin{table}[ht]
\begin{tabular}{|l|c|c|c|}
\hline
\multicolumn{1}{|c|}{Agents} & \begin{tabular}[c]{@{}c@{}}Success\\ rate (\%)\end{tabular} & \begin{tabular}[c]{@{}c@{}}Finish lap\\ time (s)\end{tabular} & \begin{tabular}[c]{@{}c@{}}Steering $S_m$\\ \end{tabular} \\ \hline
Vanilla SAC                  & 59\%                                                        & 18.12 ± 0.63     & 0.070   \\ \hline
Temporal                  & 65\%                                                        & 18.16 ± 0.62                                                          & 0.066                                                     \\ \hline
Spatial                    & 84\%                                                        & 13.26 ± 0.45                                                           & 0.059                                                     \\ \hline
I-RAS                        & 87\%                                                        & 13.34 ± 0.35                                                          & 0.059                                                  \\ \hline
\end{tabular}
\caption{Ablation study of I-RAS components by evaluating the four agents, vanilla SAC, Temporal, Spatial, and I-RAS (including both temporal and spatial smoothness).
}
\label{tab:ablation_i_ras}
\end{table}

\subsubsection{Photometric and Geometric Transformation}
\label{sec:ablation_study_spatial_component}
In this ablation experiment, we study the effectiveness of photometric and geometric transformation in spatial smoothness regularization by ablating either photometric or geometric transformation.
In addition to the agents with vanilla SAC and Spatial (the same as the ones in Table \ref{tab:ablation_i_ras}), we consider two more agents with the following models: 
\begin{inparaenum}[(a)]
    \item The agent uses photometric transformation only to generate images in spatial smoothness (denoted as Photometric).
    \item The agent uses geometric transformation only (denoted as Geometric). 
\end{inparaenum}



\autoref{tab:ablation_spatial} shows that geometry transformation improves the performances more significantly than photometric transformation. 
It is not hard to conclude this from the table in terms of any of the finish lap time, success rate, and steering smoothness value. 
A possible reason is that a vanilla agent that includes RandConv, a kind of photometric transformation method for sim-to-real transfer, also has the capability to deal with different photometric environment conditions.
Therefore, the photometric transformation takes less impact to smooth the control action, and Photometric does not improve performance significantly over vanilla SAC.
The geometry is used to simulate the possible next state from the current state by rotating, shifting, or scaling which is important to learn how to smooth the trajectory in a complex environment like an autonomous racing car.


\begin{table}[ht]
\begin{tabular}{|l|c|c|c|}
\hline
\multicolumn{1}{|c|}{Agents} & \begin{tabular}[c]{@{}c@{}}Success\\ rate (\%)\end{tabular} & \begin{tabular}[c]{@{}c@{}}Finish lap\\ time (s)\end{tabular} & \begin{tabular}[c]{@{}c@{}}Steering $S_m$\\ \end{tabular} \\ \hline
Vanilla SAC                  & 59\%          & 18.12 ± 0.63     & 0.070   \\ \hline
Photometric                  & 60\%           & 18.17 ± 0.51                 & 0.071               \\ \hline
Geometric                    & 83\%       & 13.28 ± 034         & 0.059               \\ \hline
Both (Spatial)                    & 84\%                                                        & 13.26 ± 0.45                                                           & 0.059                                                     \\ \hline
\end{tabular}
\caption{Ablation study of photometric and geometric transformation in spatial smoothness by evaluating the four agents, vanilla SAC, Photometric, Geometric, and Spatial.
}
\label{tab:ablation_spatial}
\end{table}

\section{Conclusion}
\label{sec:conclusion}

This paper presents image-based regularization for action smoothness (I-RAS) to smooth the control of autonomous miniature car racing.
The agent with I-RAS achieves a smoother trajectory compared to the agent without I-RAS, therefore improving the performance in both terms of success rate and finish lap time.
We also introduce the IR control, an adaptive regularization weight combined between normalized speed and normalized action.
The IR control allows the agent to quickly change the action in needed cases while maintaining a smooth trajectory during the run.
This is critical to improve the performance of autonomous racing.
We also demonstrated the robustness and effectiveness of our method in the AWS DeepRacer competition, a platform for an autonomous racing car. 
Our agent with I-RAS won the 2022 AWS DeepRacer Final Championship with a record of finishing a track in 13.756 seconds.

We further analyze the impact of I-RAS components by ablation study. 
It is concluded from experiments that spatial smoothness takes more impact on performance than temporal smoothness. 
This is in contrast with the low-dimensional input problem for CAPS, where temporal smoothness is the most important component.
Besides, it is also concluded from experiments that geometric transformation takes more impact on performance than photometric transformation. 
A possible reason is that photometric transformation is also included in RandConv for sim-to-real transfer.
More future work is expected to clarify these issues. 


\section*{Acknowledgement}
This research was supported in part by the National Science and Technology Council (NSTC) of the Republic of China (Taiwan) under Grants 110-2221-E-A49-067-MY3, 111-2221-E-A49-101-MY2, and 111-2634-F-A49-013. 

\bibliographystyle{plain}
\bibliography{root}

\end{document}